\documentclass[final]{cvpr}

\usepackage{times}
\usepackage{epsfig}
\usepackage{graphicx}
\usepackage{amsmath}
\usepackage{amssymb}
\usepackage{multirow}
\usepackage{booktabs}
\usepackage{color}
\usepackage{xspace}
\usepackage{bm}
\usepackage{multirow}
\usepackage{stfloats}

\usepackage[pagebackref=true,breaklinks=true,colorlinks,bookmarks=false]{hyperref}

\def\cvprPaperID{56} 
\def\confYear{CVPR 2022}

\begin{document}

\title{EPIC-KITCHENS-100 Unsupervised Domain Adaptation Challenge for Action Recognition 2022: Team HNU-FPV Technical Report}

\author{Nie Lin, Minjie Cai\\
College of Computer Science and Electronic Engineering, Hunan University\\
Hunan, China\\
{\tt\small \{nielin,caiminjie\}@hnu.edu.cn}\\
}

\maketitle

\begin{abstract}
In this report, we present the technical details of our submission to the 2022 EPIC-Kitchens Unsupervised Domain Adaptation (UDA) Challenge. Existing UDA methods align the global features extracted from the whole video clips across the source and target domains but suffer from the spatial redundancy of feature matching in video recognition. Motivated by the observation that in most cases a small image region in each video frame can be informative enough for the action recognition task, we propose to exploit informative image regions to perform efficient domain alignment. Specifically, we first use lightweight CNNs to extract the global information of the input two-stream video frames and select the informative image patches by a differentiable interpolation-based selection strategy. Then the global information from videos frames and local information from image patches are processed by an existing video adaptation method, i.e., TA3N, in order to perform feature alignment for the source domain and the target domain. Our method (without model ensemble) ranks 4th among this year's teams on the test set of EPIC-KITCHENS-100.
\end{abstract}

\section{Introduction}
With the rapid development of deep learning techniques, how to develop deep neural networks to understand human's daily interactions with surrounding environments from the first-person perspective has gained increasing interests from researchers. The EPIC-KITCHENS-100 dataset is a large video dataset of first-person perspective, and the videos record most of the common actions that would happen in a kitchen scene ~\cite{damen2020ek100}. The dataset provides fine-grained action labels, and each action is composed by a pair of verb and noun labels. In order to meet the task of EPIC-KITCHENS-100 Unsupervised Domain Adaptation (UDA) Challenge for Action Recognition, the model needs to be trained on the labeled source domain (EPIC-KITCHENS-2018) and adapted to the unlabeled target domain (EPIC-KITCHENS-100). The UDA for action recognition is more challenging than the action recognition task since the adapted model needs to overcome the domain discrepancy represented in complex video features between the source domain and the target domain. Therefore, how to effectively model the shared feature representation of the source and target domains is one of the keys to solve this challenge.

\begin{figure}
    \centering
    \includegraphics[width=\linewidth]{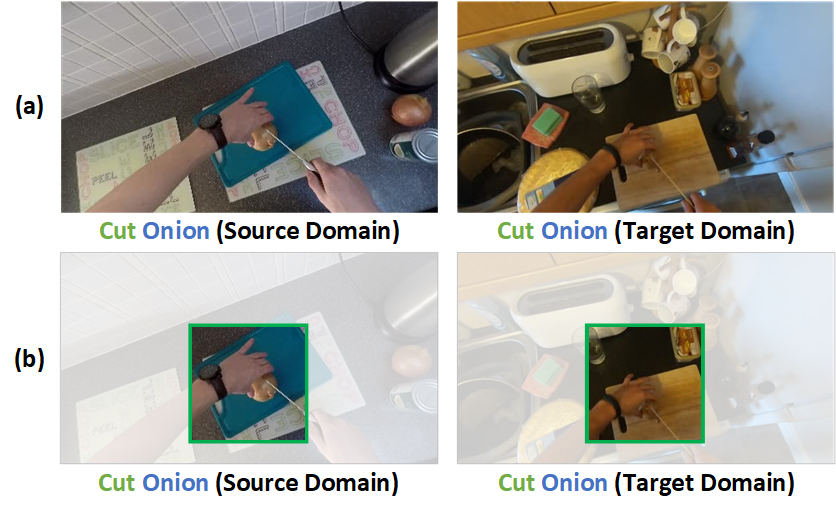}
    \caption{Illustration of fine-grained action recognition on EPIC-KITCHENS-55 (source domain) and EPIC-KITCHENS-100 (target domain). (a) Due to the differences in shooting time and indoor environment, there are many different background objects in the video clip of the same action (\eg, ``cutting onion'') between the source/target domains, which are irrelevant to the action recognition task. (b) By selecting the most informative image regions for processing, the domain discrepancy between the source domain and the target domain can be effectively reduced.}
    \label{fig:figure_1}
\end{figure}

\begin{figure*}[ht]
    \centering
    \includegraphics[width=\linewidth]{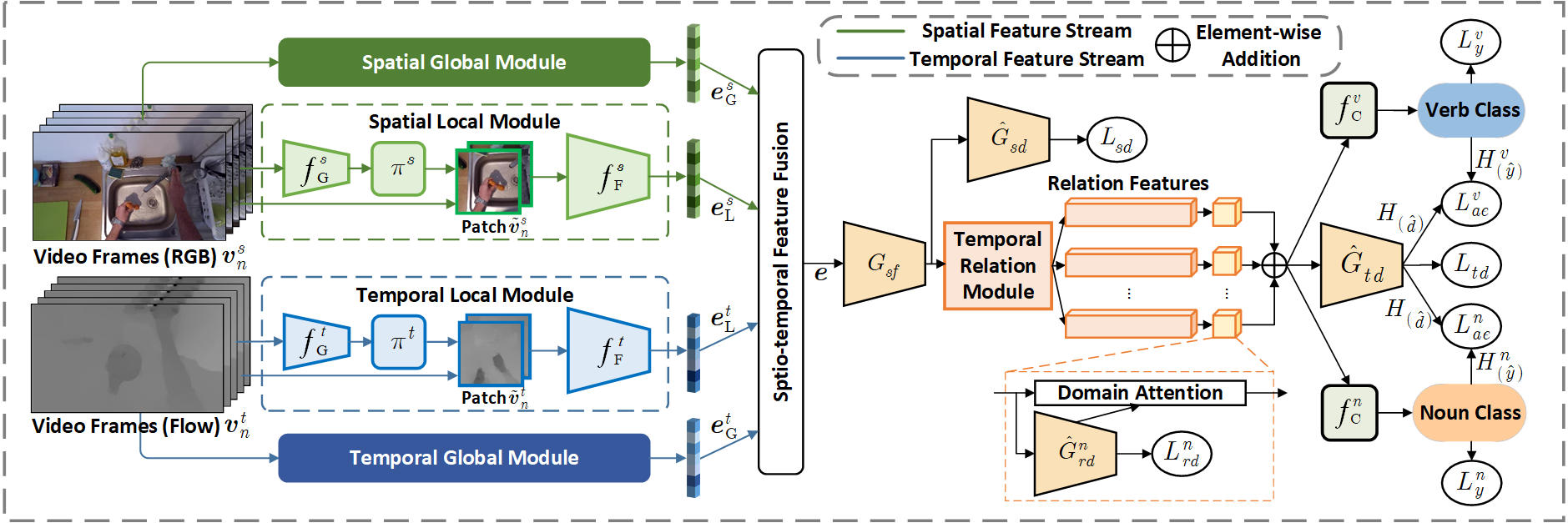}
    \caption{Overview of the proposed method. The method includes two main parts: spatio-temporal feature extraction and video domain adaptation. In spatio-temporal feature extraction, it is composed by global feature extraction branches and local feature extraction branches for both RGB and optical flow inputs. ${f}^{s}_{\text{G}}$, ${f}^{s}_{\text{F}}$ and $\pi{ }^{s}$ denote the glancer, focuser and policy networks for the spatial local module, respectively. Similar notations are used for the temporal local module. In video domain adaptation, $\hat{G}^{ }_{sd}$, $\hat{G}^{ }_{td}$ and $\hat{G}^{n}_{rd}$ denote the spatial, temporal and relation domain classifiers, respectively. ${L}^{ }_{sd}$, ${L}^{ }_{td}$ and ${L}^{n}_{rd}$ denote the spatial, temporal and relation domain classification loss. ${f}^{v}_{\text{C}}$ and ${f}^{n}_{\text{C}}$ denote the verb classifier and noun classifier. ${L}^{v}_{y}$ and ${L}^{n}_{y}$ denote the verb and noun classification loss, respectively. ${L}^{v}_{ae}$ and ${L}^{n}_{ae}$ denote the attentive entropy loss for verb and noun, respectively.}
    \label{fig:overview}
\end{figure*}

As recorded by a wearable camera from the first-person perspective, egocentric video is characterized by rapidly changing background between consecutive actions and cluttered background containing multiple objects irrelevant to the ongoing action. Furthermore, for videos in different domains, the same actions may present huge differences of image appearance, especially in the background. As a result, directly modeling shared feature representation between different domains is challenging due to spatial redundancy in the original video features. Figure~\ref{fig:figure_1} shows examples of video frames of the same action from two different domains. It can be seen that the action of ``cutting onion'' in the source domain shows quite different visual appearance compared with the target domain. One exception is the region around hands which show certain consistency between two domains. Actually, information of the verb ``cutting'' and the noun ``onion'' is fully encoded in such informative regions of video frames. 
So the challenge of action recognition in UDA lies in the frequent scene switching between each action and the difference in the background of the same action in different domains. 
Therefore, instead of straightforward domain alignment of original video features, exploiting the most informative regions in video frames for feature extraction shows a promising way of efficient domain adaptation for egocentric action recognition.

In this work, we incorporate a learning-based patch selection strategy into an existing video domain adaption framework. The patch selection strategy is implemented as a lightweight CNN and a policy network which helps locate the task-related regions and extract local features for each video frame. We consider both RGB and optical flow images as input to capture the spatial and temporal characteristic of an action. After spatial-temporal feature fusion with both global and local features, we adopt an existing video domain adaptation method TA3N~\cite{Chen_2019_ICCV} to do feature alignment for the source and target domains. The experimental results on EPIC-KITCHEN-100 demonstrate the effectiveness of the proposed method in UDA for action recognition.

\section{Method}
As an overview of our approach is described in Figure~\ref{fig:overview}, the overall model is divided into two parts. The first part of the model extracts the spatio-temporal features of the video from the input RGB frames and optical flow frames and contains both global and local branches in the process. For the local branch, inspired by the latest work in video-based action recognition~\cite{Wang_2021_ICCV, Wang_2022_CVPR}, we build a spatio-temporal local feature extraction. After extracting the global and local features of the original video, the model will fuse the spatio-temporal features extracted from different domains through spatio-temporal feature fusion. In the second part, the model is used to align the spatio-temporal features extracted from the source domain and the target domain and finally complete the action prediction of the target domain. We will introduce the above component in detail in the following sections.

\begin{table*}
\centering
\caption{The recognition performance of different models on target validation set. FeatDim: the dimension of shared features of TA3N; NumSeg: the number of input frames between the global and local branches is from left to right. The left and right side of ``+'' indicates the input into the glancer network and the focuser network. ``-'' indicates that local branches are not used for feature extraction.\\}
\label{tab:epic-val}
\begin{tabular}{c|c|c|c|c|c|c|c|c|c|c} 
\hline
\multirow{2}{*}{Method} & \multicolumn{2}{c|}{Backbone} & \multirow{2}{*}{FeatDim} & \multirow{2}{*}{NumSeg} & \multicolumn{3}{c|}{Top-1 Accuracy (\%)} & \multicolumn{3}{c}{Top-5 Accuracy (\%)}  \\ 
\cline{2-3}\cline{6-11}
                        & Global & Local                &                          &                         & Verb  & Noun  & Action                   & Verb  & Noun  & Action                   \\ 
\hline
TA3N                    & TBN    & -                    & 512                      & 6/-                       & 48.10 & 26.74 & 18.72                    & 77.98 & 47.50 & 41.87                    \\ 
\hline
TA3N                    & TBN    & -                    & 1024                     & 6/-                       & 48.28 & 27.30 & 19.25                    & 76.71 & 47.39 & 41.65                    \\ 
\hline
TA3N                    & TBN    & MN2/RN               & 1024                     & 6/4+6                     & 48.70 & 27.87 & 19.61                    & 76.18 & 48.52 & 42.01                    \\ 
\hline
TA3N                    & TBN    & MN2/RN               & 2048                     & 12/8+12                    & 49.42 & 28.33 & 20.11                    & 77.06 & 47.52 & 41.82                    \\
\hline
\end{tabular}
\end{table*}

\begin{table*}
\centering
\caption{The recognition performance of different models on target test set. All results on the test set were evaluated on the test server. Table column definitions are the same as in Table~\ref{tab:epic-val}.\\}
\label{tab:epic-test}
\begin{tabular}{c|c|c|c|c|c|c|c|c|c|c} 
\hline
\multirow{2}{*}{Method} & \multicolumn{2}{c|}{Backbone} & \multirow{2}{*}{FeatDim} & \multirow{2}{*}{NumSeg} & \multicolumn{3}{c|}{Top-1 Accuracy (\%)} & \multicolumn{3}{c}{Top-5 Accuracy (\%)}  \\ 
\cline{2-3}\cline{6-11}
                        & Global & Local                &                          &                         & Verb  & Noun  & Action                   & Verb  & Noun  & Action                   \\ 
\hline
TA3N                    & TBN    & MN2/RN               & 1024                     & 6/4+6                     & 47.71 & 27.74 & 19.41                    & 73.38 & 48.91 & 31.26                    \\ 
\hline
TA3N                    & TBN    & MN2/RN               & 2048                     & 12/8+12                    & 48.87 & 28.72 & 19.88                    & 74.61 & 49.70 & 32.32                    \\
\hline
\end{tabular}
\end{table*}

\subsection{The Spatio-temporal Feature Extraction}
Given a RGB stream of video frames $\left\{ \bm{v}^{s}_{1},\bm{v}^{s}_{2},... \right\}$ and a optical flow stream  of video frames $\left\{ \bm{v}^{t}_{1},\bm{v}^{t}_{2},... \right\}$, the model will extract the spatio-temporal features of the two different video streams. For the local feature extraction, the model takes a glance at each frame in the video with the corresponding glancer network ${f}^{ }_{\text{G}}$. Then the cheap and coarse feature will be fed into the corresponding policy network $\pi{ }^{ }$ to select the area that contributes the most to the task:
\begin{equation}
\begin{split}
    \bm{\tilde{v}}^{s}_{n} = \pi^{s}({f}^{s}_{\text{G}}(\bm{{v}}^{s}_{n})),\quad n = 1,2,...,\\
    \bm{\tilde{v}}^{t}_{n} = \pi^{t}({f}^{t}_{\text{G}}(\bm{{v}}^{t}_{n})),\quad n = 1,2,...,\\
\end{split}
\end{equation}
where $\tilde{\bm{v}}^{s}_{n}$, $\tilde{\bm{v}}^{t}_{n}$ are the selected patches of RGB video frames and optical flow video frames of the $n^{th}_{}$ frame. And the selected patches $\tilde{\bm{v}}^{s}_{n}$, $\tilde{\bm{v}}^{t}_{n}$ will be fed into the corresponding focuser network ${f}^{ }_{\text{F}}$ to extract the local feature maps $\bm{e}^{s}_{\text{L}}$, $\bm{e}^{t}_{\text{L}}$:
\begin{equation}
\begin{split}
    \bm{{e}}^{s}_{\text{L}} = {f}^{s}_{\text{F}}(\bm{\tilde{v}}^{s}_{n}),\quad n = 1,2,...,\\
    \bm{{e}}^{t}_{\text{L}} = {f}^{t}_{\text{F}}(\bm{\tilde{v}}^{t}_{n}),\quad n = 1,2,...,\\
\end{split}
\end{equation}

Finally, the global spatio-temporal features $\bm{e}^{s}_{\text{G}}$, $\bm{e}^{t}_{\text{G}}$ and audio feature $\bm{e}^{a}_{\text{G}}$ extracted from the global branch. Note that our model considers the global features corresponding to the audio modalities, which are not represented in the figure for the sake of simplicity. Then the global features are concatenated with the local spatio-temporal features $\bm{e}^{s}_{\text{L}}$, $\bm{e}^{t}_{\text{L}}$ extracted from the local branch are concatenated to serve as the input final feature $\bm{e}^{ }_{ }$ to the video domain adaptation:
\begin{equation}
    \bm{e} = Concat(\bm{e}^{s}_{\text{G}}, \bm{e}^{s}_{\text{L}}, \bm{e}^{t}_{\text{G}}, \bm{e}^{t}_{\text{L}}, \bm{e}^{a}_{\text{G}}).
\end{equation}

\subsection{The Video Domain Adaptation}
After the global and local spatio-temporal features are obtained, it will be more beneficial for the model to perform efficient domain alignment. In the video domain-adapted training of global-local features from the source and target domains, we adapt an existing video domain adaptation method for action recognition tasks, \ie, TA3N~\cite{Chen_2019_ICCV}. As shown in Figure~\ref{fig:overview}, model first aligns frame-level features from the source and target domain inputs through the adversarial discriminators $\hat{G}^{ }_{sd}$ and generates the corresponding domain loss ${L}^{ }_{sd}$. At the same time, the frame-level features of the input are modeled in the temporal relation module of TA3N, and these relation features are aggregated to obtain the video-level features. In aggregating these relational features, the domain attention mechanism is added to pay more attention to the alignment of local temporal features that have larger domain discrepancy. In the domain attention mechanism, the adversarial discriminators $\hat{G}^{n}_{rd}$ are used to align the relational features from the source and target domains, and the corresponding domain loss ${L}^{n}_{rd}$ is generated. Then, the adversarial discriminators $\hat{G}^{ }_{td}$ are also used to align the video-level features from the source and target domains, and the corresponding domain loss ${L}^{ }_{td}$ is generated. Finally, the model classifies the video-level features through two corresponding classifiers ${f}^{v}_{\text{C}}$ and ${f}^{n}_{\text{C}}$, and generates the predicted verb classification and noun classification.

\section{Experiments}


\subsection{Implementation Details}
\begin{figure}[h]
    \centering
    \includegraphics[width=\linewidth]{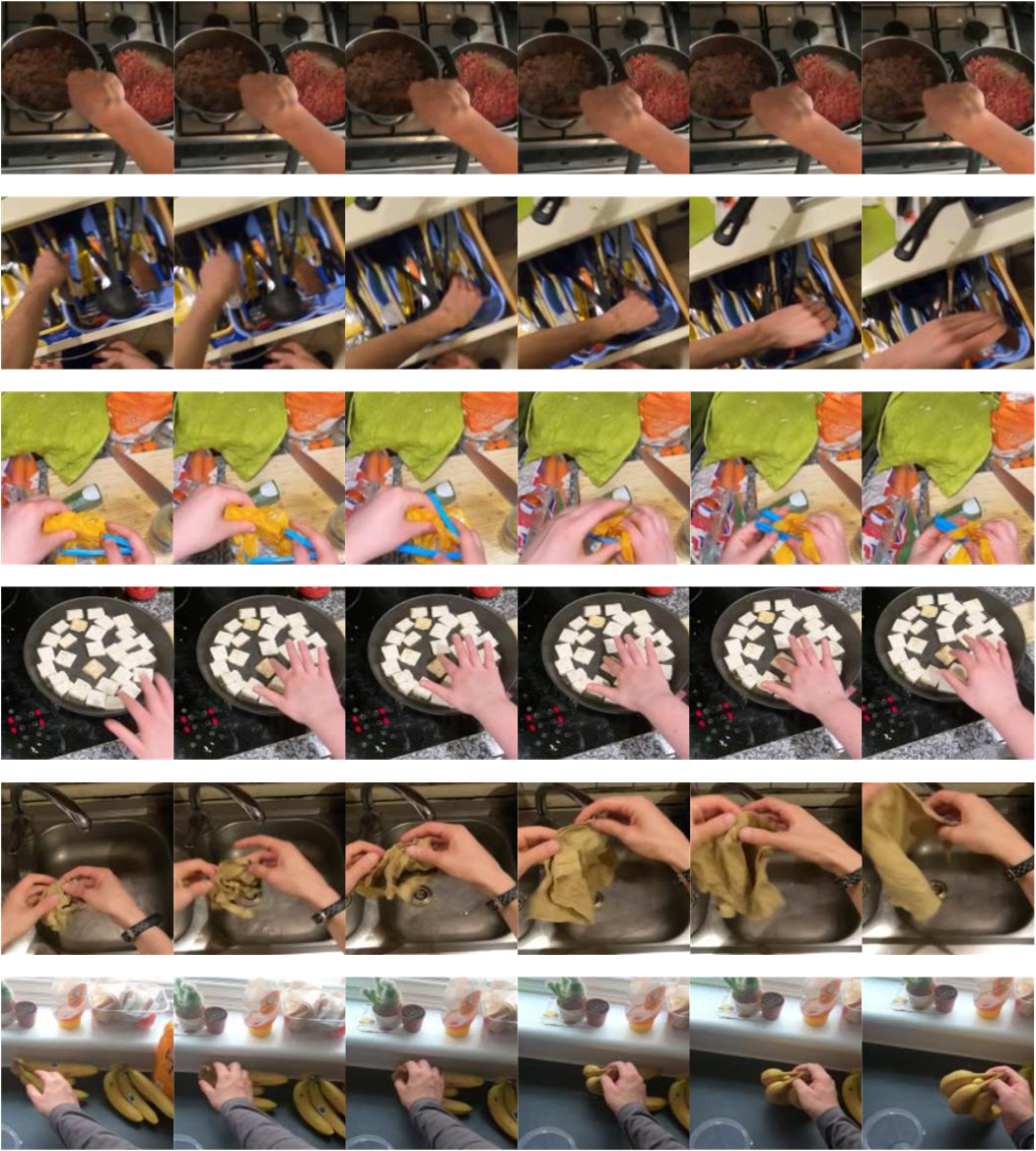}
    \caption{Visualization results of the image patches selected from the spatial local module.}
    \label{fig:figure_3}
\end{figure}

\textbf{Spatio-temporal feature extraction.} Since the network of spatial feature extraction and temporal feature extraction are the same in parameter settings, the following description will not distinguish between spatial and temporal feature extraction. For the global feature of spatio-temporal feature extraction, we use RGB, flow and audio features provided by the organizers that were extracted with Temporal Binding Network (TBN) ~\cite{Kazakos_2019_ICCV} pretrained in the source domain. And we follow the model setting in ~\cite{Wang_2022_CVPR} to extract the local feature of the spatio-temporal feature. We also adopt MobileNet-V2 (MN2)~\cite{Sandler_2018_CVPR} and ResNet-50 (RN)~\cite{He_2016_CVPR} as the glancer network ${f}^{ }_{\text{G}}$ and focuser network ${f}^{ }_{\text{F}}$, respectively. And the same policy network is used to select the image patch that contributes most to the task from the input video frames by the differentiable bilinear interpolation. The network parameters are learned with SGD optimizer with momentum 0.9 and weight decay $5 \times {10}^{-4}_{ } $. For each network of the local branches, the initial learning rates of ${f}^{ }_{\text{G}}$, ${f}^{ }_{\text{F}}$ and $\pi{ }^{ }$ are set to $0.005$, $0.01$, and $1e-4$, respectively. For the video frames that are input into the model, we adopt the same processing method as ~\cite{Wang_2022_CVPR} and set the size of the selected image patch to $176 \times 176$.

\textbf{Video domain adaptation.} After obtaining the spatio-temporal features of the source and target domains, TA3N~\cite{Chen_2019_ICCV} is used to align the input features and generate the prediction results of the model. The network parameters are also learned with SGD optimizer with momentum 0.9 and weight decay $5 \times {10}^{-4}_{ }$. During training, the parameters in the spatio-temporal feature extraction are freezed. The initial learning rate is set at 3e-3 and decayed by a factor of 0.1 at epochs 10 and 20.

\subsection{Result}

Table~\ref{tab:epic-val} shows the action recognition effect of the model on the target validation set under different input and hyper-parameter settings. The table shows that the accuracy can be improved under the same hyperparameter setting by using the local spatio-temporal branch to extract the local feature. We tried two groups of models trained under different hyper-parameters, and their performance on the target test set is shown in Table~\ref{tab:epic-test}. Our proposed method performs favorably against TA3N by $0.93\%$ in the top-1 action accuracy. In our final submission, we use RGB, Flow and Audio modalities, and the shared feature dimension of the model is set as 2048. The number of input frames of the glancer network and focuser network is set as 8 and 12, respectively. And the visualization results of the image patches selected from the test set by the proposed method are shown in Figure~\ref{fig:figure_3}. Each line shows a number of image patches selected from consecutive video frames by the spatial local module of the method. It should be noticed that the spatial local module is fixed after training with source domain data. It can be seen that the model can also be well applied to the videos of the target domain.

\section{Conclusion}
This paper presents the technical details of our solution for the EPIC-KITCHENS-100 UDA for Action Recognition Challenge. By incorporating a learning-based patch selection strategy into an existing video domain adaption
framework, the proposed method can effectively improve the domain adaptation performance of action recognition. Our work empirically verifies the importance of exploiting informative regions for egocentric videos and provides some new inspirations for domain adaptive action recognition.

{\small
	\bibliographystyle{ieee_fullname}
	\bibliography{egbib}

\begin{thebibliography}{10}\itemsep=-1pt

\bibitem{arnab2021vivit}
Anurag Arnab, Mostafa Dehghani, Georg Heigold, Chen Sun, Mario Lu{\v{c}}i{\'c},
  and Cordelia Schmid.
\newblock Vivit: A video vision transformer.
\newblock {\em arXiv preprint arXiv:2103.15691}, 2021.

\bibitem{bertasius2021timesformer}
Gedas Bertasius, Heng Wang, and Lorenzo Torresani.
\newblock Is space-time attention all you need for video understanding?
\newblock {\em arXiv preprint arXiv:2102.05095}, 2021.

\bibitem{carreira2019k700}
Joao Carreira, Eric Noland, Chloe Hillier, and Andrew Zisserman.
\newblock A short note on the kinetics-700 human action dataset.
\newblock {\em arXiv preprint arXiv:1907.06987}, 2019.

\bibitem{carreira2017i3d}
Joao Carreira and Andrew Zisserman.
\newblock Quo vadis, action recognition? a new model and the kinetics dataset.
\newblock In {\em proceedings of the IEEE Conference on Computer Vision and
  Pattern Recognition}, pages 6299--6308, 2017.

\bibitem{damen2020ek100}
Dima Damen, Hazel Doughty, Giovanni~Maria Farinella, Antonino Furnari,
  Evangelos Kazakos, Jian Ma, Davide Moltisanti, Jonathan Munro, Toby Perrett,
  Will Price, et~al.
\newblock Rescaling egocentric vision.
\newblock {\em arXiv preprint arXiv:2006.13256}, 2020.

\bibitem{dosovitskiy2020vit}
Alexey Dosovitskiy, Lucas Beyer, Alexander Kolesnikov, Dirk Weissenborn,
  Xiaohua Zhai, Thomas Unterthiner, Mostafa Dehghani, Matthias Minderer, Georg
  Heigold, Sylvain Gelly, et~al.
\newblock An image is worth 16x16 words: Transformers for image recognition at
  scale.
\newblock {\em arXiv preprint arXiv:2010.11929}, 2020.

\bibitem{feichtenhofer2019slowfast}
Christoph Feichtenhofer, Haoqi Fan, Jitendra Malik, and Kaiming He.
\newblock Slowfast networks for video recognition.
\newblock In {\em Proceedings of the IEEE/CVF International Conference on
  Computer Vision}, pages 6202--6211, 2019.

\bibitem{goyal2017ssv2}
Raghav Goyal, Samira Ebrahimi~Kahou, Vincent Michalski, Joanna Materzynska,
  Susanne Westphal, Heuna Kim, Valentin Haenel, Ingo Fruend, Peter Yianilos,
  Moritz Mueller-Freitag, et~al.
\newblock The" something something" video database for learning and evaluating
  visual common sense.
\newblock In {\em Proceedings of the IEEE International Conference on Computer
  Vision}, pages 5842--5850, 2017.

\bibitem{han2020coclr}
Tengda Han, Weidi Xie, and Andrew Zisserman.
\newblock Self-supervised co-training for video representation learning.
\newblock {\em arXiv preprint arXiv:2010.09709}, 2020.

\bibitem{huang2021mosi}
Ziyuan Huang, Shiwei Zhang, Jianwen Jiang, Mingqian Tang, Rong Jin, and Marcelo
  Ang.
\newblock Self-supervised motion learning from static images.
\newblock {\em arXiv preprint arXiv:2104.00240}, 2021.

\bibitem{kay2017k400}
Will Kay, Joao Carreira, Karen Simonyan, Brian Zhang, Chloe Hillier, Sudheendra
  Vijayanarasimhan, Fabio Viola, Tim Green, Trevor Back, Paul Natsev, et~al.
\newblock The kinetics human action video dataset.
\newblock {\em arXiv preprint arXiv:1705.06950}, 2017.

\bibitem{loshchilov2017adamw}
Ilya Loshchilov and Frank Hutter.
\newblock Decoupled weight decay regularization.
\newblock {\em arXiv preprint arXiv:1711.05101}, 2017.

\bibitem{qing2021tca}
Zhiwu Qing, Haisheng Su, Weihao Gan, Dongliang Wang, Wei Wu, Xiang Wang, Yu
  Qiao, Junjie Yan, Changxin Gao, and Nong Sang.
\newblock Temporal context aggregation network for temporal action proposal
  refinement.
\newblock {\em arXiv preprint arXiv:2103.13141}, 2021.

\bibitem{song2019tacnet}
Lin Song, Shiwei Zhang, Gang Yu, and Hongbin Sun.
\newblock Tacnet: Transition-aware context network for spatio-temporal action
  detection.
\newblock In {\em Proceedings of the IEEE/CVF Conference on Computer Vision and
  Pattern Recognition}, pages 11987--11995, 2019.

\bibitem{teed2020raft}
Zachary Teed and Jia Deng.
\newblock Raft: Recurrent all-pairs field transforms for optical flow.
\newblock In {\em European Conference on Computer Vision}, pages 402--419.
  Springer, 2020.

\bibitem{touvron2020deit}
Hugo Touvron, Matthieu Cord, Matthijs Douze, Francisco Massa, Alexandre
  Sablayrolles, and Herv{\'e} J{\'e}gou.
\newblock Training data-efficient image transformers \& distillation through
  attention.
\newblock {\em arXiv preprint arXiv:2012.12877}, 2020.

\bibitem{tran2019csn}
Du Tran, Heng Wang, Lorenzo Torresani, and Matt Feiszli.
\newblock Video classification with channel-separated convolutional networks.
\newblock In {\em Proceedings of the IEEE/CVF International Conference on
  Computer Vision}, pages 5552--5561, 2019.

\bibitem{wang2020cbr}
Xiang Wang, Baiteng Ma, Zhiwu Qing, Yongpeng Sang, Changxin Gao, Shiwei Zhang,
  and Nong Sang.
\newblock Cbr-net: Cascade boundary refinement network for action detection:
  Submission to activitynet challenge 2020 (task 1).
\newblock {\em arXiv preprint arXiv:2006.07526}, 2020.

\bibitem{wu2019lfb}
Chao-Yuan Wu, Christoph Feichtenhofer, Haoqi Fan, Kaiming He, Philipp
  Krahenbuhl, and Ross Girshick.
\newblock Long-term feature banks for detailed video understanding.
\newblock In {\em Proceedings of the IEEE/CVF Conference on Computer Vision and
  Pattern Recognition}, pages 284--293, 2019.

\bibitem{yun2019cutmix}
Sangdoo Yun, Dongyoon Han, Seong~Joon Oh, Sanghyuk Chun, Junsuk Choe, and
  Youngjoon Yoo.
\newblock Cutmix: Regularization strategy to train strong classifiers with
  localizable features.
\newblock In {\em Proceedings of the IEEE/CVF International Conference on
  Computer Vision}, pages 6023--6032, 2019.

\bibitem{zach2007tvl1}
Christopher Zach, Thomas Pock, and Horst Bischof.
\newblock A duality based approach for realtime tv-l 1 optical flow.
\newblock In {\em Joint pattern recognition symposium}, pages 214--223.
  Springer, 2007.

\bibitem{zhang2017mixup}
Hongyi Zhang, Moustapha Cisse, Yann~N Dauphin, and David Lopez-Paz.
\newblock mixup: Beyond empirical risk minimization.
\newblock {\em arXiv preprint arXiv:1710.09412}, 2017.

\bibitem{zhao2020pointtransformer}
Hengshuang Zhao, Li Jiang, Jiaya Jia, Philip Torr, and Vladlen Koltun.
\newblock Point transformer.
\newblock {\em arXiv preprint arXiv:2012.09164}, 2020.

\bibitem{zhong2020randomerasing}
Zhun Zhong, Liang Zheng, Guoliang Kang, Shaozi Li, and Yi Yang.
\newblock Random erasing data augmentation.
\newblock In {\em Proceedings of the AAAI Conference on Artificial
  Intelligence}, volume~34, pages 13001--13008, 2020.

\bibitem{zhou2021deepvit}
Daquan Zhou, Bingyi Kang, Xiaojie Jin, Linjie Yang, Xiaochen Lian, Zihang
  Jiang, Qibin Hou, and Jiashi Feng.
\newblock Deepvit: Towards deeper vision transformer.
\newblock {\em arXiv preprint arXiv:2103.11886}, 2021.

\end{thebibliography}


\begin{thebibliography}{1}\itemsep=-1pt

\bibitem{Chen_2019_ICCV}
Min-Hung Chen, Zsolt Kira, Ghassan AlRegib, Jaekwon Yoo, Ruxin Chen, and Jian
  Zheng.
\newblock Temporal attentive alignment for large-scale video domain adaptation.
\newblock In {\em Proceedings of the IEEE/CVF International Conference on
  Computer Vision}, pages 6321–6330, 2019.

\bibitem{damen2020ek100}
Dima Damen, Hazel Doughty, Giovanni~Maria Farinella, Antonino Furnari,
  Evangelos Kazakos, Jian Ma, Davide Moltisanti, Jonathan Munro, Toby Perrett,
  Will Price, et~al.
\newblock Rescaling egocentric vision.
\newblock {\em arXiv preprint arXiv:2006.13256}, 2020.

\bibitem{He_2016_CVPR}
Kaiming He, Xiangyu Zhang, Shaoqing Ren, and Jian Sun.
\newblock Deep residual learning for image recognition.
\newblock In {\em Proceedings of the IEEE/CVF Conference on Computer Vision and
  Pattern Recognition}, pages 770–778, 2016.

\bibitem{Kazakos_2019_ICCV}
Evangelos Kazakos, Arsha Nagrani, Andrew Zisserman, and Dima Damen.
\newblock Epic-fusion: Audio-visual temporal binding for egocentric action
  recognition.
\newblock In {\em Proceedings of the IEEE/CVF International Conference on
  Computer Vision}, pages 5492–5501, 2019.

\bibitem{Sandler_2018_CVPR}
Mark Sandler, Andrew Howard, Menglong Zhu, Andrey Zhmoginov, and Liang-Chieh
  Chen.
\newblock Mobilenetv2: Inverted residuals and linear bottlenecks.
\newblock In {\em Proceedings of the IEEE/CVF Conference on Computer Vision and
  Pattern Recognition}, pages 4510–4520, 2018.

\bibitem{Wang_2021_ICCV}
Yulin Wang, Zhaoxi Chen, Haojun Jiang, Shiji Song, Yizeng Han, and Gao Huang.
\newblock Adaptive focus for efficient video recognition.
\newblock In {\em Proceedings of the IEEE/CVF International Conference on
  Computer Vision}, pages 16249--16258, 2021.

\bibitem{Wang_2022_CVPR}
Yulin Wang, Yang Yue, Yuanze Lin, Haojun Jiang, Zihang Lai, Victor Kulikov,
  Nikita Orlov, Humphrey Shi, and Gao Huang.
\newblock Adafocus v2: End-to-end training of spatial dynamic networks for
  video recognition.
\newblock In {\em Proceedings of the IEEE/CVF Conference on Computer Vision and
  Pattern Recognition}, pages 20062--20072, 2022.

\end{thebibliography}
}

\end{document}